%% file: main.tex
\documentclass{article} 
\usepackage{main,times}

\input{math_commands.tex}

\usepackage{hyperref}
\usepackage{url}
\usepackage{graphicx}
\usepackage{booktabs}
\usepackage{xcolor}
\graphicspath{{./figs/}}

\newcommand{\figfile}[1]{\IfFileExists{figs/#1}{\includegraphics{#1}}{\framebox[0.95\linewidth]{\rule{0pt}{1.2in}\texttt{missing figure: #1}}}}
\newcommand{\figfilefit}[1]{\IfFileExists{figs/#1}{\includegraphics[width=\linewidth]{#1}}{\framebox[0.95\linewidth]{\rule{0pt}{1.2in}\texttt{missing figure: #1}}}}

\title{Two Routes to the Middle: Placement Search and Brain Readouts Converge on Where Continual Learners Should Specialize}

\author{Yuan Huang
\quad Zihan Chen
\quad Runbin Zhang
\quad Hongwei Ding
\quad Changzeng Fu
\quad Shiqi Zhao \thanks{Corresponding author.}\\
Northeastern University \\
}

\iclrfinalcopy 
\begin{document}

\maketitle

\begin{abstract}
Continual learners that keep a task-specific adapter in every block of a pre-trained vision transformer accumulate storage linearly with the number of tasks; keeping task-specific adapters in only a few blocks curbs this growth but raises the question of where to place them. We investigate this question from two perspectives. Algorithmically, training all contiguous four-block placements yields an inverted U: final accuracy peaks at intermediate depth and varies by up to 3.5 percentage points (pp), while inexpensive criteria based on weight spectra or activation statistics favor the deepest blocks. From neuroscience, the hierarchical organization and intermediate-stage plasticity of the visual cortex motivate us to ask whether a measurement taken outside the learner can guide layer specialization without placement search. LS-B observes the first tasks through a frozen fMRI encoding model of twelve human visual areas and commits task-specific capacity once to the blocks whose readouts vary most across tasks relative to their stable structure. Across three ViT-B/16 backbones, LS-B yields stable, backbone-specific allocations. On the two backbones with placement search, AugReg and iBOT, the selected blocks overlap the intermediate-depth region identified by search. Under matched storage and observation budgets, the selected blocks outperform the shallowest and deepest four-block configurations. On Split ImageNet-R, LS-B uses 60\% of full-BiLoRA adapter storage while remaining within 1.5~pp of its final accuracy. The allocation requires no labels or backpropagation, adds under 0.6\% runtime, and exhibits backbone-specific cortical signatures.
\end{abstract}

\section{Introduction}
\label{sec:intro}

\begin{figure}[t]
  \centering
  \resizebox{0.93\linewidth}{!}{\figfile{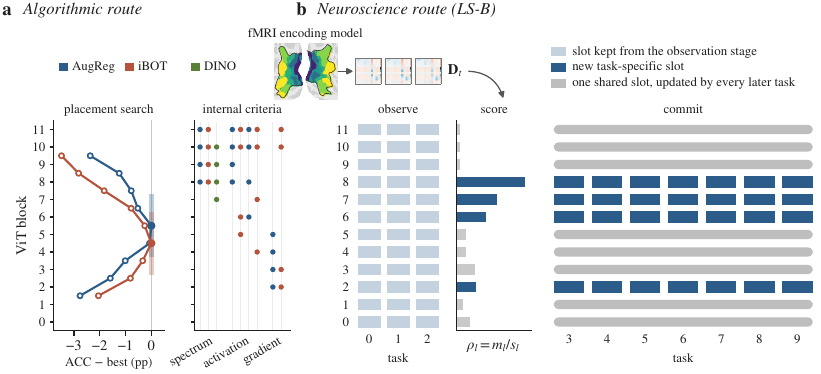}}
  \caption{Two routes to the intermediate blocks (Split ImageNet-R). \textbf{(a)}~Placement search with no observation (48 slots; seeds 0--2) peaks at blocks 4--7 (AugReg) and 3--6 (iBOT); dots show blocks selected by internal criteria. \textbf{(b)}~LS-B: a frozen fMRI encoding model turns each of the first $K=3$ tasks into a readout $\mathbf{D}_t$; $\rho_l=m_l/s_l$ commits $k=4$ blocks and the other eight share one slot. Example: AugReg seed~0 commits blocks 2, 6, 7, 8 (72 of 120 slots); allocations vary across seeds.}
  \label{fig:overview}
\end{figure}

Pre-trained vision transformers with parameter-efficient modules are a strong basis for rehearsal-free class-incremental learning \citep{l2p,dualprompt,inflora,bilora}. If the learner keeps a task-specific module in every block, as BiLoRA does \citep{bilora}, the parameters it retains grow linearly with the task sequence. Keeping task-specific capacity in only $k$ of the $L$ blocks and sharing one parameter-efficient adaptation module elsewhere (hereafter, \emph{adapter}) curbs this growth \citep{cllora}, but raises a structural question: \emph{which blocks should keep task-specific capacity?} On 10-task Split ImageNet-R, moving a four-block task-specific window along a ViT-B/16 changes final accuracy by up to 2.8 and 3.5~pp on two backbones at identical storage (Figure~\ref{fig:overview}a).

The \emph{algorithmic route} to an answer looks inside the learner. Placement can be searched on validation data \citep{dualprompt,surgicalft}; CL-LoRA, which fixes the shared/task-specific boundary, notes that ``the optimal position could be determined through validation on the first few tasks'' \citep[p.~30540]{cllora}. On our benchmark, search works, since every contiguous four-block placement trained to completion traces an inverted U with an intermediate optimum, but it costs one full continual-learning run per candidate and seed. Cheaper criteria read a per-layer signal from the learner's parameters, activations or gradients \citep{hidellava,ckpdals,dmole,rainbowprompt,treelora}; we find that weight-spectrum and activation criteria favor the deepest blocks and miss the optimum.

The \emph{neuroscience route} looks outside the learner. Deep vision networks parallel the ventral stream, with early layers predicting early visual cortex and later layers higher areas \citep{yamins2014performance,guclu2015deep}, and layer-resolved encoding models assign each cortical location to the network depth that predicts it best \citep{braindecodes}. In adult visual cortex, perceptual learning modifies responses in the intermediate area V4 \citep{yang2004perceptual,raiguel2006learning}, while plasticity of early visual cortex is constrained by critical periods \citep{hensch2005critical}, which have an analogue in deep networks \citep{achille2019critical}. These findings leave open where task-specific capacity belongs and motivate a question: can an independently fitted neural measurement of how each block maps onto visual cortex locate the blocks that need task-specific capacity without labels or placement search?

\begin{figure}[t]
  \centering
  \figfile{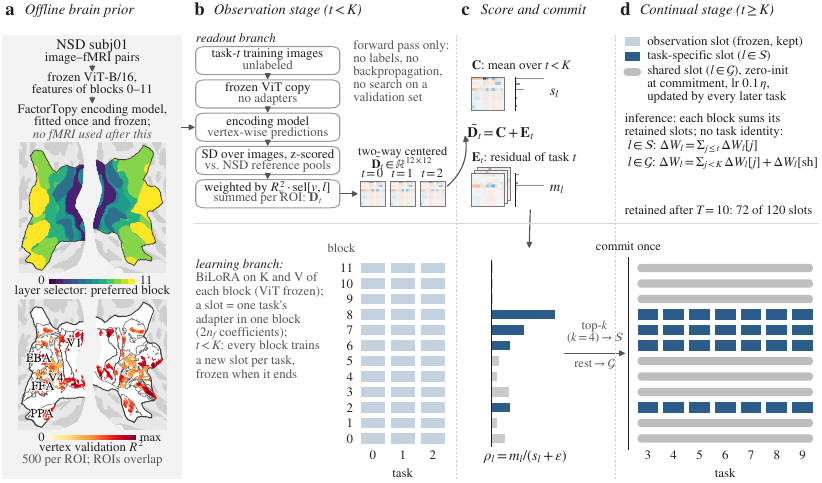}
  \caption{LS-B end to end (AugReg, Split ImageNet-R). \textbf{(a)}~Offline, once per backbone: an encoding model fitted to NSD subj01 maps frozen block features to cortex (flatmaps: preferred block per vertex; $R^2$ of the retained vertices). \textbf{(b)}~Observation ($t<K=3$): every block trains a task-specific slot, while unlabeled task images pass forward through an adapter-free copy and the encoding model to give $\mathbf{D}_t$. \textbf{(c)}~$\widetilde{\mathbf{D}}_t=\mathbf{C}+\mathbf{E}_t$ yields $s_l$, $m_l$ and $\rho_l$; the top $k=4$ blocks are committed once. \textbf{(d)}~Later tasks add slots only to committed blocks; the others update one shared slot. Matrices: diagnostic readout; slots: seed~0 (chosen by index; committed blocks vary across seeds).}
  \label{fig:pipeline}
\end{figure}

We answer with LS-B (Layer Specialization from external Brain readouts), an observe-then-commit rule (Figure~\ref{fig:pipeline}). During the first $K$ tasks every block is task-specific, while a frozen fMRI encoding model of twelve human visual areas reads an adapter-free copy of the backbone on each task's unlabeled images. LS-B then scores each block by how much its readout varies across tasks relative to its stable part, commits task-specific capacity once to the $k$ highest-scoring blocks, and lets the other blocks share one continually updated adapter.

The two routes meet (Figure~\ref{fig:overview}): the readout is not concentrated in the deepest blocks and commits blocks in the intermediate band found by placement search. The readout recovers the intermediate-region preference without recovering the search optimum: the selected sets overlap with the best-performing window by one to three blocks out of four, and its ranking of candidate placements shows no detectable correlation with their accuracy. Our contributions are (i) a measurement of the algorithmic route: placement follows an inverted U with an intermediate optimum that spectral and activation criteria miss (\S\ref{sec:algo}); (ii) a method from the neuroscience route: a frozen, layer-resolved fMRI readout turned into a one-time shared/task-specific partition without labels, backpropagation or search (\S\ref{sec:method}); and (iii) evidence of convergence and of its price: gains over both extremes at matched storage and observation, 60\% of full-adapter storage within 1.5~pp of its accuracy, and backbone-specific cortical signatures (\S\ref{sec:neuro}--\S\ref{sec:cortex}).

\section{Related Work}
\label{sec:related}

\paragraph{Parameter-efficient continual learning.} With pre-trained backbones, continual learners \citep{cl_survey2} adapt per task through adapters \citep{houlsby_adapter}, prompts \citep{l2p,dualprompt} or low-rank updates \citep{lora,inflora,bilora}, often separating shared from task-specific knowledge \citep{dualprompt,cllora}. Adapting only the first session can already provide a strong continual-learning baseline \citep{fsa,zhou2025revisiting}; RanPAC likewise adopts this strategy as its first phase \citep{ranpac}. The observation stage of LS-B is related to this form of early adaptation; after observation, LS-B also fixes the layer-wise capacity allocation. A similar observe-then-commit idea appears in adaptive-rank fine-tuning \citep{adalora}. LoRA-drop keeps LoRA in the layers with the most important LoRA outputs and shares one LoRA across the remaining layers \citep{loradrop}. Automatic per-layer capacity allocation is also a central objective of D-MoLE \citep{dmole}.

\paragraph{Where to adapt.} Layer-wise allocation is decided by fixed rules \citep{cllora}, by search or validation \citep{dualprompt,surgicalft}, by empirical comparison of placements \citep{aesp}, or by learner-derived signals: representational similarity \citep{hidellava}, sensitivity statistics \citep{ckpdals}, gradient magnitudes \citep{dmole}, learned gates \citep{rainbowprompt} or gradient similarity \citep{treelora}, in networks whose features become more task-specific with depth \citep{yosinski2014transferable}. All of these read their evidence from the learning system; LS-B obtains its allocation signal through a frozen neural encoding model fitted independently of the continual-learning process.

\paragraph{Brains and networks.} Networks are compared with neural data along the ventral stream \citep{yamins2014performance,guclu2015deep,brainscore}, and layer-resolved encoding models map voxels onto layers \citep{braindecodes}; LS-B uses such a model as a frozen instrument. Neural data have regularized models \citep{brainreg,federer}, with part of the benefit matched by non-neural data of similar statistics \citep{federer}, fine-tuned language and speech models \citep{brainbias,braintuning,brainwavlm,braintuning2}, and informed the choice of intervention layers \citep{nari_narf}. Brain-inspired continual learners import complementary memories \citep{fearnet}, expert gating \citep{expertgate}, internal replay \citep{bir} or random-expanded routing \citep{flyprompt}; LS-B uses a layer-resolved neural measurement directly in the layer-allocation rule.

\section{Method}
\label{sec:method}

\subsection{Setting and host adapter}
\label{sec:host}

We study class-incremental learning over $T$ tasks without task identity at test time, on a frozen ViT-B/16 \citep{vit} with $L=12$ blocks. LS-B builds on BiLoRA \citep{bilora} as the host adapter and a FactorTopy encoding model \citep{braindecodes} as the neural readout, and specifies how the readout determines a shared/task-specific partition of the blocks and how that partition is maintained across tasks. In its released implementation, BiLoRA adapts the key and value projections of every block. For task $\tau$, the update of one projection $\mathbf{W}\in\mathbb{R}^{d\times d}$ ($d=768$) is
\[
\Delta\mathbf{W}^{(\tau)}=\alpha\,\mathrm{Re}\,\mathcal{F}^{-1}\big(P_\tau(\mathbf{c}^{(\tau)})\big),
\]
where $P_\tau$ places the $n_f=3000$ trainable coefficients $\mathbf{c}^{(\tau)}$ at positions of a $d\times d$ spectrum that are fixed for each task index, $\mathcal{F}^{-1}$ is the two-dimensional inverse discrete Fourier transform and $\alpha=300$ is the released scale; a block computes its keys and values with $\mathbf{W}$ plus the sum of the updates it retains. We call what one task keeps in one block a \emph{task--layer slot} (6{,}000 parameters). A finished task's slot is frozen and keeps contributing to the forward pass; full BiLoRA retains $LT=120$ slots (720{,}000 parameters) after $T=10$ tasks.

\subsection{Brain readout}
\label{sec:readout}

\paragraph{Encoding model.} We fit a FactorTopy encoding model on image--fMRI pairs of subject subj01 of the Natural Scenes Dataset (NSD; \citealp{nsd}). It predicts cortical responses from the frozen backbone's block features through a per-vertex layer-selection weight over the $L$ blocks, is fitted once per backbone, and is then frozen; no fMRI data are used during continual learning. We read it out in $R=12$ visual regions (V1, V2, V3, V4, EBA, FBA, OFA, FFA, OPA, PPA, OWFA and VWFA), keeping in each region the 500 vertices with the highest validation $R^2$ (details and Figure~\ref{fig:atlas} in Appendix~\ref{app:fmri}). 
The regions overlap anatomically: 13.7--14.2\% of the retained vertices belong to more than one region, concentrated in a few pairs such as FBA and FFA, whose rows of $\mathbf{D}_t$ below are close to duplicates; we therefore treat the regions as overlapping groups of vertices.

\paragraph{Task-evoked readout.} At the start of each observation task $t$, up to ten batches of 128 of the task's unlabeled training images pass through a separate, adapter-free frozen copy of the backbone and then through the encoding model. 
Let $\sigma_t[n,l]$ be the across-image standard deviation of the response that the encoding model predicts for retained vertex $n$ from block $l$ alone. The same statistic on ten reference pools of NSD training images, each as large as the task sample, has mean $\mu[n,l]$ and standard deviation $\nu[n,l]$ over pools, and $z_t[n,l]=(\sigma_t[n,l]-\mu[n,l])/\nu[n,l]$. With $\mathrm{sel}[n,l]$ the vertex's layer-selection weight and $q_{r,n}$ its validation $R^2$, clipped at zero and normalized to sum to one over the retained vertices of region $r$, the readout $\mathbf{D}_t\in\mathbb{R}^{R\times L}$ is
\[
\mathbf{D}_t[r,l]=\sum_{n\in r} q_{r,n}\,\mathrm{sel}[n,l]\,z_t[n,l];
\]
entry $(r,l)$ measures how strongly the task's images modulate the responses that block $l$ predicts in region $r$. Because the copy carries no adapters, adaptation from earlier tasks does not enter $\mathbf{D}_t$.

\subsection{Layer specialization score}
\label{sec:score}

Raw readout values are dominated by region- and block-specific scales, so LS-B compares interactions: it removes the additive main effects of regions and blocks from each $\mathbf{D}_t$ by two-way centering, $\widetilde{\mathbf{D}}_t[r,l]=\mathbf{D}_t[r,l]-\bar{\mathbf{D}}_t[r,\cdot]-\bar{\mathbf{D}}_t[\cdot,l]+\bar{\mathbf{D}}_t[\cdot,\cdot]$ (bars denote means over the dotted indices), and splits the observation window into a common component and per-task residuals:
\[
\mathbf{C}=\frac{1}{K}\sum_{t=0}^{K-1}\widetilde{\mathbf{D}}_t,
\;\;
\mathbf{E}_t=\widetilde{\mathbf{D}}_t-\mathbf{C},
\;\;
m_l=\frac{1}{K}\sum_{t=0}^{K-1}\sum_{r=1}^{R}\mathbf{E}_t[r,l]^2,
\;\;
s_l=\sum_{r=1}^{R}\mathbf{C}[r,l]^2,
\;\;
\rho_l=\frac{m_l}{s_l+\epsilon},
\]
with $\epsilon=10^{-12}$ guarding against division by zero. The score asks how large a block's task-to-task deviation is \emph{relative to} the part of its brain-aligned structure that does not move across tasks: a block driven strongly but consistently across tasks therefore receives a low score and is treated by LS-B as a candidate for sharing (Figure~\ref{fig:readout} works through the score on real readouts). The score requires $K\geq2$: with one observed task the residual vanishes, $\rho\equiv0$, and the rule commits the $k$ shallowest blocks, a configuration we use only as an observation-only control.

\subsection{Observe, then commit}
\label{sec:commit}

For the first $K$ tasks, all $L$ blocks are trained with task-specific slots as in full BiLoRA; these slots are frozen, keep contributing to the forward pass and are counted in the budget. After task $K-1$, LS-B assigns the $k$ blocks with the highest $\rho_l$ to the task-specific set $\mathcal{S}$ and the rest to the shared set $\mathcal{G}$. This partition is committed once and kept fixed for all subsequent tasks. From task $K$ on, every block in $\mathcal{S}$ creates, trains and freezes a new slot for each task. Blocks in $\mathcal{G}$ retain their $K$ frozen observation-stage slots and add one persistent shared slot, which is updated by all subsequent tasks.
Writing $\Delta\mathbf{W}_l^{(\tau)}$ for block $l$'s slot of task $\tau$ and $\Delta\mathbf{W}_l^{\mathrm{sh}}$ for its shared slot, block $l$ uses after task $t\geq K$, for each of its two projections,
\[
\Delta\mathbf{W}_l=\sum_{\tau=0}^{K-1}\Delta\mathbf{W}_l^{(\tau)}+
\begin{cases}\sum_{\tau=K}^{t}\Delta\mathbf{W}_l^{(\tau)}, & l\in\mathcal{S},\\ \Delta\mathbf{W}_l^{\mathrm{sh}}, & l\in\mathcal{G}.\end{cases}
\]
While task $t$ is trained with a cross-entropy loss over its own classes, only the classifier, $\Delta\mathbf{W}_l^{(t)}$ for $l\in\mathcal{S}$ and $\Delta\mathbf{W}_l^{\mathrm{sh}}$ for $l\in\mathcal{G}$ receive gradients (every block trains $\Delta\mathbf{W}_l^{(t)}$ while $t<K$). The shared slot is zero-initialized at commitment, so attaching it leaves the network's function unchanged, and it is trained at $\eta_{\mathrm{sh}}=0.1\,\eta$ so that the state written by all remaining tasks is not dominated by the most recent one. At inference, each block sums its retained slots; no task identity is needed. After $T$ tasks LS-B retains $LK+k(T-K)+(L-k)$ slots, i.e.\ $2n_f[L(K+1)+k(T-K-1)]$ parameters: with $K=3$, $k=4$ and $T=10$, 72 slots or $432{,}000$ parameters, 60\% of full BiLoRA; the ratio approaches $k/L$ as $T$ grows. Classification heads are identical across methods and excluded.

\subsection{Implementation details and cost}
\label{sec:cost}

\paragraph{Inputs.} The commitment depends only on the unlabeled training images of the first $K$ tasks, a frozen copy of the pre-trained backbone and the frozen encoding model; it uses no labels, gradients or held-out data and none of the learner's trained parameters. Normalizing by $s_l$ matters: on the diagnostic readouts, block~11 carries the largest shared energy on every backbone and would be selected by $m_l$ alone, while the ratio $\rho_l$ does not select it (Figure~\ref{fig:readout}); no main-text LS-B run commits it (0 of 40).

\paragraph{Cost and budget.} Selection (three readouts and one ranking on $12\times12$ matrices) takes 40--46~s of a 2.1--2.3~h ten-task run on one RTX 3090, and no readout is taken after commitment; the storage saving does not shorten training, since our implementation sums all retained slots in each forward pass. At the defaults, LS-B's 72 slots are 36 observation-stage slots, 28 slots of the four committed blocks over the remaining seven tasks and 8 shared slots; a static allocation of the same width retains 48, so LS-B's 24 extra slots are the observation-stage slots of its eight shared blocks.

\section{Experimental Setup}
\label{sec:setup}

\paragraph{Benchmark and training.} We use Split ImageNet-R \citep{imagenet_r,dualprompt}: 200 classes in $T=10$ tasks of 20 disjoint classes under the class-incremental protocol. Our frozen, per-class stratified 80/20 split differs from BiLoRA's, so absolute accuracies are not comparable with published numbers; the class order is the same in every run. ACC is the final accuracy of a single 200-way classifier on all test images, without task identity, logit masking or exemplars. The ViT-B/16 backbone is initialized from AugReg (supervised, ImageNet-21k; \citealp{augreg}), iBOT \citep{ibot} or DINO \citep{dino} (self-supervised, ImageNet-1k). We keep the host adapter's recipe and hold out no validation data (Appendix~\ref{app:setup}); defaults are $k=4$ and $K=3$. All main-text experiments use this benchmark; Appendix~\ref{app:supp} reports registered supplementary runs on Split CIFAR-100, Split CUB-200 and 20-task Split ImageNet-R ($n=3$), with mixed results.

\paragraph{Configurations} (Table~\ref{tab:arms}). \emph{Placement search}: the nine windows of four contiguous task-specific blocks, 0--3 to 8--11, without observation (48 slots). \emph{Internal criteria} (48 slots): the heavy-tailed weight spectrum ($\alpha$-Hill), a D-MoLE-style gradient norm \citep{dmole}, and two activation criteria that apply our score to activations, with random or principal-component feature groups in place of the regions. \emph{Same-width controls} (48 slots): four random blocks per run, and an equal partition of the readout's vertices along its layer axis. \emph{Matched-observation controls} (72 slots): LS-B's observation stage followed by committing blocks 0--3 or 8--11. \emph{References}: all blocks shared, shallow-first prefixes of $l\in\{2,4,6,8,10\}$ blocks, the deepest ten blocks, and full BiLoRA. \emph{Sweeps}: $k\in\{1,\dots,8\}$ at $K=3$, and $K\in\{1,2\}$ at $k=4$.

\paragraph{Capacity and statistics.} Same-width controls retain 48 slots against 72 for LS-B and 120 for full BiLoRA; we state the slot counts with every comparison and test placement only against the matched-observation controls. Seed subsets are fixed by index: three-seed comparisons use seeds 0--2, and configurations with more runs (LS-B and random allocation: 16/16/8 on AugReg/iBOT/DINO; criteria, equal partition and the deepest four: 8) are also reported at full $n$. Differences are paired by seed. The comparisons with random allocation and the activation criteria were registered before their runs as one-sided paired $t$-tests at a Bonferroni-corrected level of .003125 (DINO: .025; the other same-width controls: descriptive); the matched-observation readouts were registered as descriptive quantities; the sweeps and decompositions are exploratory. We make no significance claims at $n=3$ and use a registered run-to-run gap of 0.42~pp, an upper bound on seed noise, as a yardstick for magnitudes only. Main-text results rest on 374 ten-task runs, 13 of them bitwise reproductions (Appendix~\ref{app:stats}).

\section{Results and Discussion}
\label{sec:results}

\subsection{The algorithmic route: search finds the middle, internal criteria miss it}
\label{sec:algo}

\begin{figure}[t]
  \centering
  \figfile{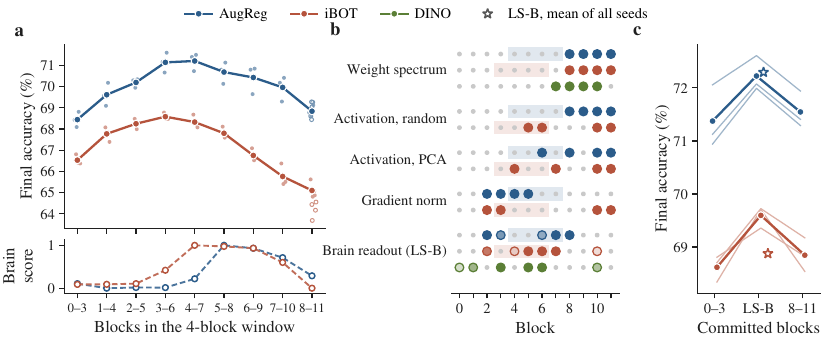}
  \caption{Both routes lead to the intermediate blocks. \textbf{(a)}~Top: final accuracy of a four-block window at 48 slots without observation (large: mean of seeds 0--2; small: single seeds; open: five further seeds of window 8--11). Window 0--3 extends the eight registered windows. Bottom: readout window score $\sum_{l\in w}\rho_l$ ($K=3$ diagnostic readout, min--max scaled). \textbf{(b)}~Blocks selected by each criterion; LS-B: fraction of seeds committing each block (solid: all seeds); boxes: best window of (a). \textbf{(c)}~Matched storage and observation ($K=3$, 72 slots): only the committed blocks differ. Thin lines: single seeds; star: LS-B mean over 16 seeds.}
  \label{fig:routes}
\end{figure}

Figure~\ref{fig:routes}a slides the four-block window along depth at 48 slots. Accuracy follows an inverted U, peaking at blocks 4--7 on AugReg (71.20\%) and 3--6 on iBOT (68.58\%). Over the eight registered windows (1--4 to 8--11) the range is 2.37 and 3.48~pp, against 0.49~pp expected from seed noise, and the deepest window is the worst on both backbones (eight-seed means 68.89\% and 64.65\%). The shallowest window, 0--3 (not tested), is also poor (68.44\% and 66.53\%). Search thus locates an intermediate optimum, on the slice of contiguous windows (the best of all 495 four-block allocations is unknown) and at the cost of 27 runs per backbone.

Internal criteria do not (Figure~\ref{fig:routes}b). The weight-spectrum criterion is almost a depth ranking (Spearman correlation with depth $+0.87$ to $+1.00$) and selects blocks 8--11 (7--10 on DINO); the activation criteria select sets with mean depth 8.0--9.5 (blocks 0--11), where search finds the lowest accuracy. The gradient-norm criterion is the exception (blocks 2--5 on AugReg, $\{2,3,10,11\}$ on iBOT): the depth bias belongs to spectral and activation statistics.

\subsection{The neuroscience route: the readout lands in the intermediate band}
\label{sec:neuro}

The readout score is not a depth proxy: its Spearman correlation with depth across blocks is $-0.21$ to $+0.19$ (descriptive). Every AugReg seed commits blocks $\{2,7,8\}$, every iBOT seed $\{5,6,7\}$ and every DINO seed $\{3,5,6\}$; with the varying fourth block, the committed sets have mean depth 4.56--5.33 (Figure~\ref{fig:allocation}). They avoid the deep end chosen by spectral and activation criteria and fall in the band around the search optima, whose windows are centered at 5.5 and 4.5.

The readout approaches the optimum without reproducing it. Its committed sets share one or two of four blocks with the best AugReg window and two or three with the best iBOT window, and its window score peaks one window deeper than accuracy (Figure~\ref{fig:routes}a, bottom). Its ranking of placements does not track accuracy: its rank correlation with accuracy is $+0.10$ and $+0.38$ over the eight windows ($p=0.84$ and $0.36$) and $+0.09$ and $-0.02$ over 32 random allocations per backbone (non-significant; these reuse the runs of the random control and are not an independent line of evidence). On iBOT the score is essentially nonzero only for blocks 6 and 7, a property registered before this analysis and not offered as an explanation of the null result.

\subsection{Where the routes meet: matched storage and observation}
\label{sec:converge}

Same-width comparisons cannot isolate placement, because observation adds 24 slots. Figure~\ref{fig:routes}c therefore fixes storage and observation (72 slots, $K=3$) and varies only the committed blocks. The readout's blocks beat the shallowest four by 0.85 and 0.98~pp and the deepest four by 0.68 and 0.75~pp on AugReg and iBOT (seed-paired, $n=3$, positive for every seed). Against LS-B's 16-seed mean, the margins are 0.92 and 0.26~pp over the shallowest and 0.75 and 0.03~pp over the deepest blocks: iBOT's seeds 0--2 are its best three of 16, and its margin over the deepest blocks does not survive this change of reference. The two extremes differ by 0.17 and 0.23~pp. At matched capacity, the intermediate blocks are better on AugReg and at least as good as either extreme on iBOT.

The same controls split observe-then-commit into its two parts (a descriptive decomposition). With blocks 0--3 committed, observing $K=3$ tasks adds 2.93 and 2.09~pp (48 to 72 slots), with diminishing returns (the first observed task, 8 of the 24 slots, adds 1.66 and 1.21~pp); replacing 0--3 by the readout's blocks adds 0.85 and 0.98~pp ($K=1$ commits 0--3 because $\rho\equiv0$, and so isolates observation). Observation supplies most of LS-B's gain over a static allocation of the same width; the readout decides where the remaining task-specific capacity goes.

\subsection{Storage--accuracy trade-off}
\label{sec:frontier}

\begin{figure}[t]
  \centering
  \figfile{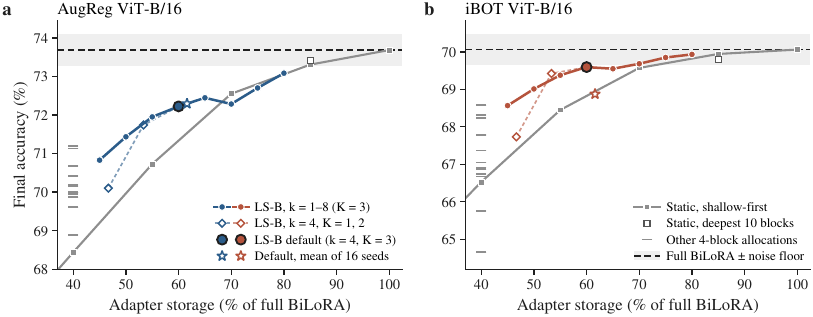}
  \caption{Storage--accuracy trade-off (\textbf{a}~AugReg, \textbf{b}~iBOT); storage: retained slots relative to full BiLoRA. LS-B with $k=1$--8 ($K=3$) lies above the static shallow-first allocations for $k\leq5$ and within 0.42~pp of them for $k\geq6$; the default uses 60\%. Stars: the default over 16 seeds. Dashes at 40\%: all other four-block allocations (all runs). Dashed line and band: full BiLoRA $\pm$ 0.42~pp.}
  \label{fig:frontier}
\end{figure}

\begin{figure}[t]
  \centering
  \begin{minipage}[t]{0.485\linewidth}
    \centering
    \figfile{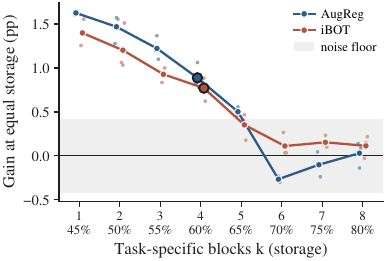}
    \caption{Gain over a storage-matched interpolation of static shallow-first allocations (seed-paired, $n=3$; large markers: means). At $k=4$ the gain against the 16-seed mean is 0.96 and 0.05~pp. Grey: the 0.42~pp yardstick.}
    \label{fig:matched_storage}
  \end{minipage}\hfill
  \begin{minipage}[t]{0.485\linewidth}
    \centering
    \figfile{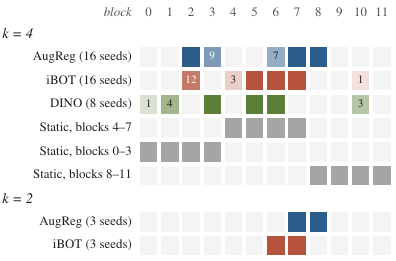}
    \caption{Committed blocks ($K=3$; shade: fraction of seeds, partial cells give seed counts). At $k=4$ every seed commits $\{2,7,8\}$ (AugReg), $\{5,6,7\}$ (iBOT) and $\{3,5,6\}$ (DINO); at $k=2$, $\{7,8\}$ and $\{6,7\}$.}
    \label{fig:allocation}
  \end{minipage}
\end{figure}

With $k=4$ and $K=3$, LS-B keeps 72 of 120 slots (60\%) and reaches 72.22\%, 69.60\% and 64.37\% on AugReg, iBOT and DINO (seeds 0--2; 72.29\%, 68.88\% and 64.39\% over 16, 16 and 8 seeds), within 1.5~pp of full BiLoRA (73.69\%, 70.06\% and 65.13\%; Figure~\ref{fig:frontier}). For $k\leq5$, LS-B lies above the static shallow-first allocations: its storage-matched gain over their interpolation is 1.63 and 1.40~pp at $k=1$ and 0.89 and 0.77~pp at $k=4$ (seed-paired; 0.96 and 0.05~pp against the 16-seed mean), and within the 0.42~pp yardstick for $k\geq6$ (Figure~\ref{fig:matched_storage}). The interpolation is not run, and the gain does not extend to the best 48-slot windows (71.20\% and 68.58\%, against 70.83\% and 68.57\% for LS-B at $k=1$ with 54 slots). The default lies above the convex hull of the shallow-first allocations and the $k=2$ point on seeds 0--2 (not over all seeds on iBOT), and on the hull of all tested configurations on seeds 0--2 (below it over all iBOT seeds; Appendix~\ref{app:results}).

Against same-width controls (48 slots, against 72 for LS-B), LS-B is 2.32 and 1.83~pp more accurate than random allocation on AugReg and iBOT ($n=16$ pairs, all positive; $p=2.8\times10^{-12}$ and $1.7\times10^{-9}$), 1.83~pp on DINO ($n=8$, $p=7.3\times10^{-6}$), and 1.6--2.3~pp more accurate than the gradient-norm, PCA-activation and equal-partition controls ($n=8$). A prefix analysis attributes these margins mainly to observation: they arise during observation and shrink after commitment, although the final margin over random allocation is 2.11 and 2.09 times what dilution of the observation-stage advantage alone would leave (Appendix~\ref{app:prefix}). These comparisons are therefore not evidence that the readout picks better blocks; \S\ref{sec:converge} is the test of placement. Selection costs 0.53--0.57\% of runtime; the storage saving does not shorten training.

\subsection{Mapping the committed blocks onto visual cortex}
\label{sec:cortex}

\begin{figure}[t]
  \centering
  \figfile{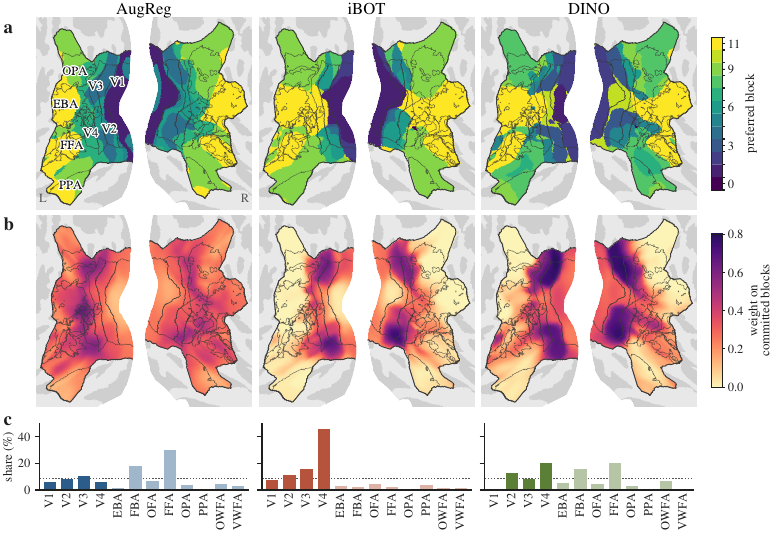}
  \caption{Committed blocks on visual cortex (flattened fsaverage, NSD visual mask; lines: region outlines). \textbf{(a)}~Layer map of the frozen encoding model: each vertex colored by the block with the largest layer-selection weight (a possibly weak peak). \textbf{(b)}~Footprint of LS-B's allocation, $\sum_l f_l\,\mathrm{sel}[v,l]$, with $f_l$ the fraction of seeds (16, 16, 8) committing block $l$; uniform: $4/12$. \textbf{(c)}~Share of the task residual of the four top-ranked blocks of the $K=3$ diagnostic readout per region (dark: V1--V4; dotted: $1/12$). Panels (b) and (c) use different committed sets (seed frequencies vs.\ diagnostic top four; overlap 3/4, 4/4, 4/4). Overlapping vertices count in both regions.}
  \label{fig:brainmap}
\end{figure}

The readout lives in cortical coordinates, so allocations map onto the brain (Figure~\ref{fig:brainmap}). The frozen encoding model orders cortex by depth (a): V1 is best predicted by shallow blocks (modal block 1 or 2), the category-selective areas EBA, FBA, FFA and VWFA by the last block and the scene areas PPA and OPA by blocks 8--9, broadly consistent with the correspondence between depth and the ventral hierarchy \citep{yamins2014performance,guclu2015deep}; V2--V4 lie in between, non-monotonically (per-region profiles in Appendix~\ref{app:fmri}, Figure~\ref{fig:atlas}). Weighting the map by how often each block is committed (b) gives LS-B's cortical footprint: on all three backbones it lies below the uniform $4/12$ in V1 and in PPA, OPA and EBA (only marginally for DINO in V1, 0.32, and for AugReg in EBA, 0.33) and peaks in between, at locations that differ between backbones. This is a post-hoc description, partly implied by (a) and the intermediate depth of the committed blocks. The task residual of the top-ranked blocks (c) differs by backbone: V1--V4 carry 31--79\% of it, against an equal share of 33\%; it concentrates in V4 on iBOT, splits between V2/V4 and FFA/FBA on DINO, and lies mostly in FFA and FBA on AugReg, whose layer selection is diffuse (median per-vertex maximum weight 0.34, lowest of the three). Pre-training thus changes both which blocks are committed and the cortical composition of the signal that commits them (task maps: Figure~\ref{fig:taskmaps}). Taken together, this is a coarse correspondence rather than a recovery of the search optimum: the readout identifies the same intermediate band but does not rank candidate placements. We therefore treat it as a structural prior, not a surrogate validation objective. The matched-observation controls further separate this placement effect from the larger gain contributed by the observation stage.

\section{Conclusion}
\label{sec:conclusion}
In this study, we introduced LS-B, a method for layer specialization in continual learning based on external brain readouts. Placement search finds an intermediate optimum for task-specific capacity at the price of training every candidate, and cheap internal statistics favor the deepest blocks. A frozen fMRI encoding readout taken during a short observation stage lands in the same intermediate band and, at matched capacity, commits blocks that outperform both extremes; LS-B turns it into a one-time partition that retains 60\% of full-adapter storage within 1.5~pp of its accuracy on Split ImageNet-R. Measurements made outside a learner can thus serve as priors for architectural decisions otherwise settled by search or by the learner's own statistics.

\clearpage
\subsection*{AI use statement}
Generative AI tools assisted with language editing, figure preparation, and software tasks including implementation, code review and repository maintenance. The research questions, literature selection, experimental design and deployment, the analyses and their interpretation were carried out by the authors.

\subsection*{Ethics statement}
This work uses the Natural Scenes Dataset, collected with informed consent and ethics approval by its authors \citep{nsd}; we use only released, de-identified data of one subject to fit an encoding model and collect no new human data. The method reduces the storage cost of continual learning; we do not foresee negative societal impacts beyond those of image classification in general.

\subsection*{Reproducibility statement}
Section~\ref{sec:method} specifies the readout, the score and the parameter lifecycles; Section~\ref{sec:setup} and Appendices~\ref{app:setup}--\ref{app:stats} specify the benchmark, the training recipe, every configuration with its blocks, slots and runs, and the seed and statistics protocol. The code and a deployment guide are available at \href{https://github.com/yuan4629/LS-B}{reproducibility repository}.

\bibliography{main}
\bibliographystyle{main}

\appendix

\section{Limitations.} The readout approaches the search optimum but does not recover the best window or rank placements: we claim that it lands in the right region at the cost of three forward-only readouts, not that it finds the optimum. Most of LS-B's advantage over same-width static allocations comes from observation, and the placement test rests on three seeds per configuration, with one margin (iBOT against the deepest blocks) depending on the seed reference. All main-text results come from one benchmark and class order, one host adapter, one architecture and one fMRI subject; the neuroscience findings are motivation by analogy; the regions overlap; and at $k=4$ the readout can inject at most $\log_2\binom{12}{4}\approx8.95$ bits into the allocation.

\section{Setup details}
\label{app:setup}

\paragraph{Split and metric.} The train/test partition is stratified per class at 80/20 (24{,}002/5{,}998 images, against 24{,}000/6{,}000 in BiLoRA's split), generated once with a fixed seed by a script included in our code. Classes are ordered by the dataset's index and cut into ten consecutive groups of twenty; the reported variability therefore reflects training noise, not sensitivity to task order. ACC after the last task averages the per-task test accuracies of the single 200-way classifier weighted by the number of test images, which equals pooled accuracy over all 200 classes.

\paragraph{Training recipe.} We keep the training configuration released with the host adapter's code for CIFAR-100 unchanged: Adam with $\beta=(0.9,0.999)$ and no weight decay; learning rate $5\times10^{-4}$ for task-specific slots and $0.1\times$ that rate for shared slots; the host's cosine schedule; batch size 128; 40 epochs on the first task and 20 on each later task; fp32; random-resized crops to $224\times224$ and horizontal flips for training, resizing to 256 and center cropping for evaluation. No validation split is held out, and no hyper-parameter, epoch budget or checkpoint is selected on the benchmark; the only quantities LS-B estimates from data are the readout statistics, from the unlabeled training images of the observation tasks.

\paragraph{Readout.} Per task, the readout uses up to ten batches of 128 training images. For task $t$, block $l$ and retained vertex $n$, let $z_t[n,l]$ be the across-image standard deviation of the response predicted from block $l$, standardized against ten reference pools of NSD images of the same size. With $q_n$ a weight derived from the vertex's validation $R^2$ and $\mathrm{sel}[n,l]$ the encoding model's layer-selection weight, $\mathbf{D}_t[r,l]=\sum_{n\in r} q_n\,\mathrm{sel}[n,l]\,z_t[n,l]$.

\paragraph{Compute.} Runs use one NVIDIA RTX 3090 each. A ten-task run of the default LS-B configuration takes on average 2.26, 2.27 and 2.12~h on AugReg, iBOT and DINO, of which selection (three readouts and one ranking) takes about 46, 46 and 40~s. The storage saving does not shorten training, since our implementation sums all retained slots in each forward pass.

\section{Allocation inventory}
\label{app:arms}

\begin{table}[h]
\caption{Configurations on Split ImageNet-R. Blocks are indexed $0\ldots11$; ``Slots'' is the number of task--layer slots retained after $T=10$ tasks ($12+9k$ without observation, $48+6k$ for LS-B with $K=3$); one slot holds 6{,}000 parameters. Runs: AugReg/iBOT/DINO; ``--'': not run. $^{a}$On DINO only $k\in\{2,4\}$ and $K\in\{1,3\}$. $^{b}$At $K=1$ the score is identically zero and the rule commits blocks 0--3. $^{c}$On DINO only $l\in\{4,6,8\}$. $^{d}$Coincides with window 2--5: the same experiment. $^{e}$Coincides with the deepest four blocks; no separate runs. $^{f}$Plus 16 further independent draws per backbone on AugReg and iBOT.}
\label{tab:arms}
\begin{center}
\small
\begin{tabular}{llccc}
\toprule
Configuration & Blocks (AugReg / iBOT) & $K$ & Slots & Runs \\
\midrule
LS-B ($k=4$, $K=3$) & per run & 3 & 72 & 16/16/8 \\
LS-B, $k\in\{1,2,3,5,6,7,8\}$ & per run & 3 & $48{+}6k$ & 3/3/3$^{a}$ \\
LS-B, $K\in\{1,2\}$, $k=4$ & per run$^{b}$ & 1, 2 & 56, 64 & 3/3/3$^{a}$ \\
Matched observation, commit 0--3 & 0--3 & 3 & 72 & 3/3/-- \\
Matched observation, commit 8--11 & 8--11 & 3 & 72 & 3/3/-- \\
\midrule
Window $w\ldots w{+}3$, $w=1,\dots,7$ & $w\ldots w{+}3$ & 0 & 48 & 3/3/-- \\
Deepest four (window 8--11) & 8--11 & 0 & 48 & 8/8/-- \\
Random four & per run & 0 & 48 & 16/16/8$^{f}$ \\
Equal partition (readout layer axis) & 2,3,7,8 / 0,4,6,7 & 0 & 48 & 8/8/-- \\
Weight spectrum ($\alpha$-Hill) & 8--11$^{e}$ & 0 & 48 & -- \\
Gradient norm (D-MoLE-style) & 2,3,4,5$^{d}$ / 2,3,10,11 & 0 & 48 & 8/8/-- \\
Activation, random-12 & 8,9,10,11$^{e}$ / 5,6,10,11 & 0 & 48 & --/8/-- \\
Activation, PCA-12 & 6,8,10,11 / 4,7,10,11 & 0 & 48 & 8/8/-- \\
\midrule
All shared & -- & 0 & 12 & 3/3/3 \\
Shallow-first prefix of $l$ blocks & $0\ldots l{-}1$ & 0 & $12{+}9l$ & 3/3/3$^{c}$ \\
Deepest ten & 2--11 & 0 & 102 & 3/3/-- \\
All task-specific (full BiLoRA) & 0--11 & 0 & 120 & 3/3/3 \\
\bottomrule
\end{tabular}
\end{center}
\end{table}

LS-B allocations are computed per run and differ between seeds: on AugReg $\{2,3,7,8\}$ (9 seeds) and $\{2,6,7,8\}$ (7); on iBOT $\{2,5,6,7\}$ (12), $\{4,5,6,7\}$ (3) and $\{5,6,7,10\}$ (1); on DINO $\{1,3,5,6\}$ (4), $\{3,5,6,10\}$ (3) and $\{0,3,5,6\}$ (1). No single allocation is ``the'' LS-B allocation. The window 0--3 is the shallow-first prefix of four blocks. The equal-partition control shares its AugReg allocation with 9 of the 16 LS-B seeds, so on those seeds the two differ only in the observation stage.

\section{The fMRI readout in detail}
\label{app:fmri}

The encoding model is the FactorTopy model of \citet{braindecodes}, fitted separately for each backbone on NSD subject subj01 \citep{nsd}. Because vertices are retained per region and NSD's region definitions overlap, the $12\times500$ retained positions correspond to 5{,}246 distinct vertices on AugReg (722, or 13.8\%, in at least two regions), 5{,}214 on iBOT (739, 14.2\%) and 5{,}255 on DINO (719, 13.7\%). The overlap is concentrated (FBA$\cap$FFA, OFA$\cap$OWFA, V4$\cap$OFA), while V1, V2, V3, EBA, OPA and PPA share no retained vertices; the rows of $\mathbf{D}_t$ for FBA and FFA are therefore close to duplicates, and the effective number of independent regions is below twelve. All allocations in this paper were computed with this property; none was recomputed.

\begin{figure}[h]
  \centering
  \figfilefit{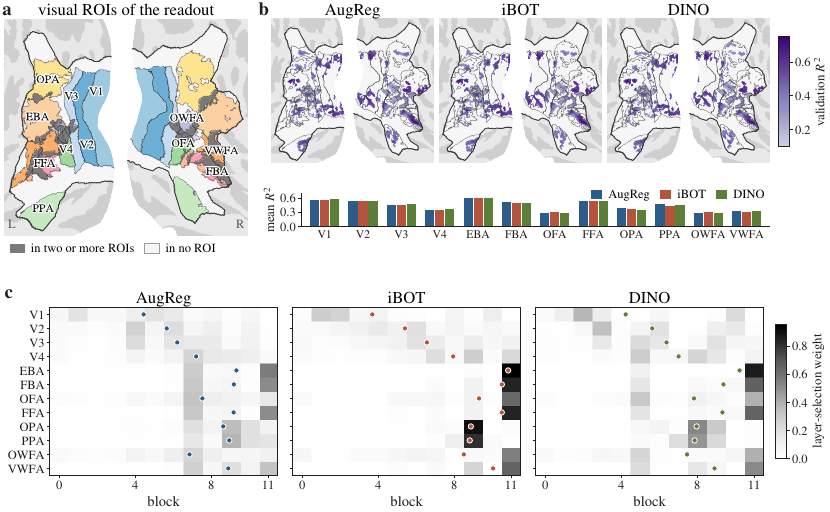}
  \caption{The frozen fMRI encoding model behind the readout (NSD subj01, flattened cortex). \textbf{(a)}~The twelve visual regions; vertices in two or more regions are grey. \textbf{(b)}~Retained vertices (500 best-fitting per region) colored by validation $R^2$, and mean $R^2$ per region. \textbf{(c)}~Weights with which the readout aggregates each region over the blocks ($R^2$-weighted mean layer selector; rows sum to one). Dots mark the weighted mean block, which rises from V1 to V4 and is largest in EBA, FBA and FFA on every backbone.}
  \label{fig:atlas}
\end{figure}

\begin{figure}[h]
  \centering
  \figfilefit{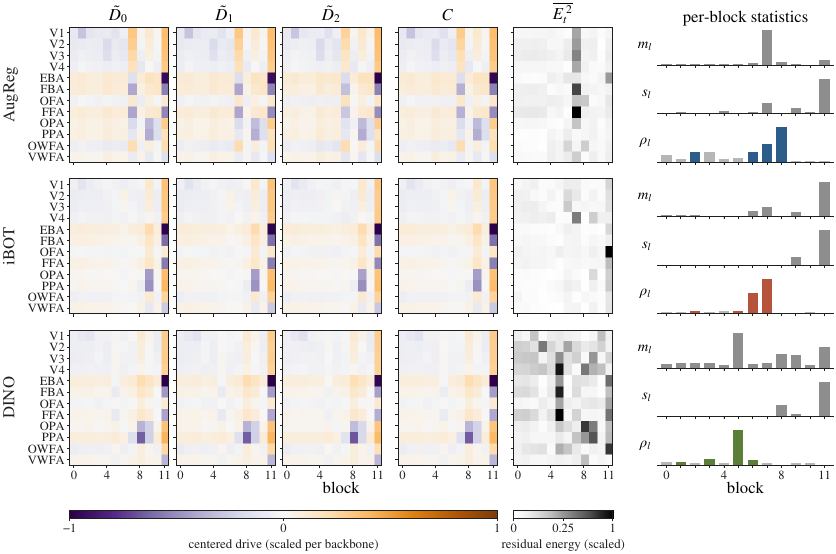}
  \caption{Worked example of the specialization score on the diagnostic readout of the three observation tasks. Rows: backbones. Columns 1--3: two-way-centered readouts $\widetilde{\mathbf{D}}_t$ (12 regions $\times$ 12 blocks); column 4: their mean $\mathbf{C}$; column 5: residual energy averaged over tasks (square-root scale). Colors are scaled per backbone. Right: $m_l$, $s_l$ and $\rho_l$; colored bars mark the four top-ranked blocks ($\{2,6,7,8\}$, $\{2,5,6,7\}$, $\{1,3,5,6\}$). Block~11 carries the largest shared energy on every backbone and would be selected by $m_l$ alone; the ratio does not select it. The task residual is 5--7\% of the centered readout. This diagnostic readout is not the per-run readout; committed sets vary across seeds (Table~\ref{tab:arms}).}
  \label{fig:readout}
\end{figure}

\begin{figure}[h]
  \centering
  \figfilefit{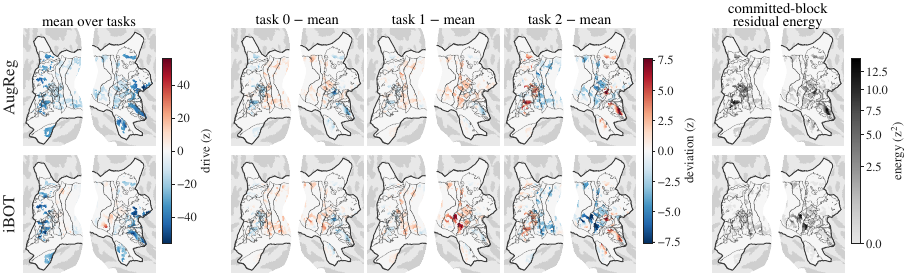}
  \caption{Task-evoked cortical readout during observation (retained vertices; AugReg and iBOT, whose vertex-level readouts were archived). Column 1: the selector-weighted, reference-standardized dispersion $\sum_l \mathrm{sel}[n,l]\,z_t[n,l]$ averaged over the three observation tasks (negative: less dispersion than on NSD images). Columns 2--4: deviation of each task from that mean. Column 5: residual energy on the four top-ranked blocks (square-root scale). These maps precede region aggregation and centering; most of the pattern is shared across tasks (pairwise correlations 0.93--0.998).}
  \label{fig:taskmaps}
\end{figure}

Figure~\ref{fig:atlas} shows the regions, the fit of the encoding model and its layer profile per region. Averaged with the readout's own weights, the preferred block rises from V1 to V4 on all three backbones and is deepest in EBA, FBA and FFA; this weighted mean differs from the per-vertex mode used in Figure~\ref{fig:brainmap}a, under which V2--V4 are not monotonic. Figure~\ref{fig:atlas}b shows the archived $R^2$ of the retained vertices only: a whole-mask $R^2$ map cannot be reproduced because the split used to validate the encoding model was not seeded. Figure~\ref{fig:readout} works through the score on the diagnostic readout of the three observation tasks, and Figure~\ref{fig:taskmaps} shows the vertex-level task signal before aggregation. Values in these figures other than the overlap counts of AugReg and iBOT and the registered scores are recomputed descriptions.

\section{Additional results}
\label{app:results}

\begin{table}[h]
\caption{LS-B operating points and static references (ACC in \%, seeds 0--2; storage relative to full BiLoRA). ``Gain'': storage-matched gain over the interpolation of static shallow-first allocations (pp, seed-paired). Default over all seeds: 72.29 / 68.88 / 64.39.}
\label{tab:sweep}
\begin{center}
\small
\begin{tabular}{lcccccc}
\toprule
Configuration & Slots & Storage & AugReg & iBOT & DINO & Gain (AugReg / iBOT) \\
\midrule
All shared & 12 & 10\% & 64.95 & 62.67 & 58.62 & -- \\
Shallow-first, $l=4$ & 48 & 40\% & 68.44 & 66.53 & 62.40 & -- \\
Best window (4--7 / 3--6) & 48 & 40\% & 71.20 & 68.58 & -- & -- \\
LS-B $k=1$ & 54 & 45\% & 70.83 & 68.57 & -- & $+1.63$ / $+1.40$ \\
LS-B $k=4$, $K=1$ & 56 & 47\% & 70.10 & 67.73 & 63.43 & $+0.64$ / $+0.35$ \\
LS-B $k=2$ & 60 & 50\% & 71.43 & 69.01 & 63.97 & $+1.47$ / $+1.20$ \\
LS-B $k=4$, $K=2$ & 64 & 53\% & 71.74 & 69.42 & -- & $+1.27$ / $+1.19$ \\
LS-B $k=3$ & 66 & 55\% & 71.95 & 69.38 & -- & $+1.22$ / $+0.93$ \\
Shallow-first, $l=6$ & 66 & 55\% & 70.73 & 68.45 & 63.90 & -- \\
LS-B $k=4$ (default) & 72 & 60\% & 72.22 & 69.60 & 64.37 & $+0.89$ / $+0.77$ \\
LS-B $k=5$ & 78 & 65\% & 72.45 & 69.55 & -- & $+0.50$ / $+0.35$ \\
LS-B $k=6$ & 84 & 70\% & 72.29 & 69.68 & -- & $-0.27$ / $+0.11$ \\
Shallow-first, $l=8$ & 84 & 70\% & 72.55 & 69.57 & 64.45 & -- \\
LS-B $k=7$ & 90 & 75\% & 72.70 & 69.85 & -- & $-0.10$ / $+0.15$ \\
LS-B $k=8$ & 96 & 80\% & 73.09 & 69.93 & -- & $+0.03$ / $+0.11$ \\
Deepest ten & 102 & 85\% & 73.41 & 69.80 & -- & -- \\
Full BiLoRA & 120 & 100\% & 73.69 & 70.06 & 65.13 & -- \\
\bottomrule
\end{tabular}
\end{center}
\end{table}

\paragraph{Convex hulls.} On seeds 0--2, the default point lies above the upper convex hull of the static shallow-first allocations (including the all-shared and full endpoints) and the $k=2$ operating point by 0.23, 0.30 and 0.16~pp on AugReg, iBOT and DINO; over all seeds, by 0.30 and 0.18~pp on AugReg and DINO, and 0.42~pp below on iBOT. Against the hull of all tested configurations, including windows and criteria, the margin is 0.03 and 0.04~pp on seeds 0--2 and 0.10 and $-0.68$~pp over all seeds.

\begin{figure}[h]
  \centering
  \figfile{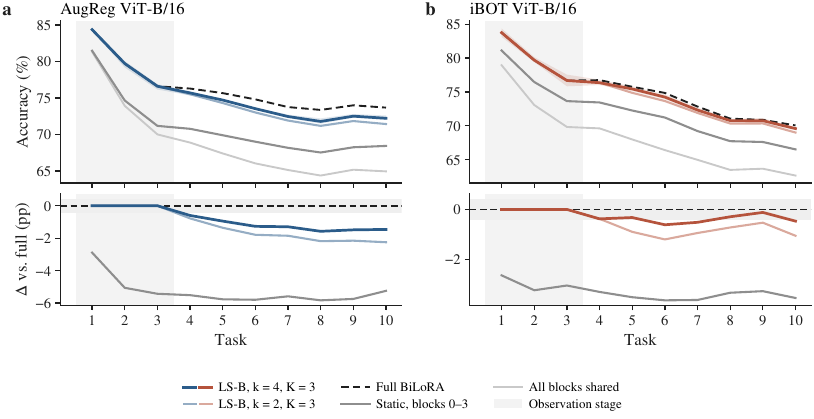}
  \caption{Accuracy after each task (\textbf{a}~AugReg, \textbf{b}~iBOT). Top: accuracy on all classes seen so far; bottom: difference from full BiLoRA; grey: observation stage. After the last task, LS-B with $k=4$ is 1.46 and 0.47~pp below full BiLoRA, with $k=2$ 2.25 and 1.05~pp, and the static allocation of blocks 0--3 5.25 and 3.53~pp.}
  \label{fig:task_curves}
\end{figure}

\subsection{Same-width controls and prefix analysis}
\label{app:prefix}

Seed-paired differences between LS-B (72 slots) and same-width controls (48 slots): random allocation $+2.32$ and $+1.83$~pp on AugReg and iBOT ($n=16$) and $+1.83$~pp on DINO ($n=8$); gradient norm $+2.00$ and $+2.29$~pp, PCA activation $+2.16$ and $+2.10$~pp, random-12 activation $+3.26$ and $+2.05$~pp (on AugReg identical to the deepest four) and equal partition $+2.15$ and $+1.60$~pp ($n=8$ each).

For a same-width control, let $d_t$ be the seed-paired difference between LS-B and the control after task $t$. We report its value when observation ends (``carry''), its change over the tasks after commitment (``late'') and its final value. Against window 4--7 (seeds 0--2), carry, late and final are 2.15, $-1.13$ and 1.02~pp on AugReg and 2.30, $-1.02$ and 1.27~pp on iBOT; against random allocation ($n=16$), 3.66, $-1.34$ and 2.32~pp and 2.91, $-1.09$ and 1.83~pp. Under the registered decision rule, a non-positive late change attributes the margin to the observation stage. The two differences are defined on different task sets (three and ten tasks), so the late change is a descriptive attribution, not a variance decomposition. The final margin over random allocation is nevertheless 2.11 and 2.09 times the value that pure dilution of the carry would leave ($0.3\times$ carry, since three of ten tasks are observed). On iBOT, two of the three LS-B seeds commit exactly blocks 4--7, so two thirds of the weight of the iBOT comparison with that window comes from pairs that differ only in the observation stage.

\section{Additional datasets and a longer task sequence}
\label{app:supp}

\begin{figure}[h]
  \centering
  \figfilefit{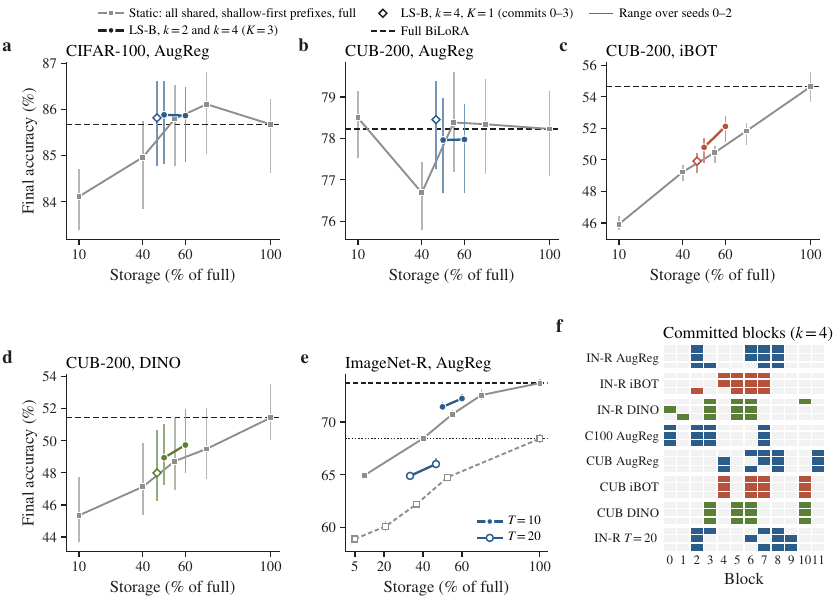}
  \caption{Supplementary conditions (seeds 0--2; bars: range over the three seeds). \textbf{(a--d)}~Final accuracy against storage relative to full BiLoRA on CIFAR-100 (AugReg) and CUB-200 (AugReg, iBOT, DINO). Gray: all shared, shallow-first prefixes of 4, 6 and 8 blocks, and full BiLoRA (also the dashed line); colored: LS-B with $k=2$ and $k=4$ ($K=3$), and LS-B with $k=4$ and $K=1$ (diamond), which commits blocks 0--3 and so adds only observation. \textbf{(e)}~ImageNet-R with AugReg at $T=10$ (filled) and $T=20$ (open; prefixes of 2, 4 and 6 blocks); dashed and dotted lines: full BiLoRA at $T=10$ and $T=20$. \textbf{(f)}~Blocks committed by LS-B ($k=4$, $K=3$) for seeds 0--2 of every condition; on ImageNet-R ($T=10$) these are the default configuration's seeds 0--2.}
  \label{fig:supp}
\end{figure}

\paragraph{Conditions.} After the main experiments we ran a supplement registered before its runs, in four conditions: Split CIFAR-100 \citep{krizhevsky2009learning} with AugReg (ten tasks of ten classes; the dataset's 50{,}000/10{,}000 split), Split CUB-200 \citep{wah2011caltech} with AugReg, iBOT and DINO (ten tasks of twenty classes; the 9{,}430/2{,}358 split distributed with the code of \citealp{zhou2025revisiting}), and Split ImageNet-R with $T=20$ tasks of ten classes (AugReg). Classes follow the dataset's index; the training configuration, the encoding model and the readout are those of the main experiments, and each dataset uses the host's data pipeline (CIFAR-100 images are upsampled to $224\times224$). Each condition trains, with seeds 0--2, all blocks shared, the shallow-first prefixes of 4, 6 and 8 blocks (2, 4 and 6 at $T=20$), LS-B with $k=4$ and $K=1$ (which commits blocks 0--3 because $\rho\equiv0$, and so adds only observation; not run at $T=20$), LS-B with $k=2$ and $k=4$ at $K=3$, and full BiLoRA. The operating points were fixed on ImageNet-R before these runs and were not re-selected, and the registered iBOT$\times$CIFAR-100 condition was not run (Appendix~\ref{app:stats}). With three seeds per configuration we report seed ranges and make no significance claims; the 0.42~pp yardstick was measured on ImageNet-R and is not applied here. Tables~\ref{tab:supp}--\ref{tab:supp_order} and Figure~\ref{fig:supp} report every configuration of every condition.

\begin{table}[h]
\caption{Supplementary ten-task conditions (ACC in \%, mean of seeds 0--2; in parentheses: range over the three seeds). Slots are retained task--layer slots after $T=10$ tasks (full BiLoRA: 120); LS-B with $K=1$ commits blocks 0--3 and so adds only observation. Bottom: registered readouts for LS-B with $k=4$ / $k=2$ ($K=3$) in pp: mean ACC minus the upper convex hull of the other seven configurations at the same slot count; seed-paired gain over the same-seed interpolation of the static shallow-first allocations; seed-paired difference from full BiLoRA. $n=3$ throughout; no significance claims.}
\label{tab:supp}
\begin{center}
\small
\setlength{\tabcolsep}{4pt}
\begin{tabular}{lccccc}
\toprule
Configuration & Slots & CIFAR-100 & CUB-200 & CUB-200 & CUB-200 \\
 & & AugReg & AugReg & iBOT & DINO \\
\midrule
All shared & 12 & 84.11\,(1.32) & 78.50\,(1.61) & 45.93\,(0.89) & 45.35\,(4.07) \\
Shallow-first, $l=4$ & 48 & 84.96\,(1.89) & 76.69\,(1.65) & 49.24\,(1.02) & 47.13\,(4.50) \\
LS-B $k=4$, $K=1$ & 56 & 85.82\,(1.85) & 78.46\,(2.12) & 49.92\,(1.31) & 47.99\,(4.37) \\
LS-B $k=2$, $K=3$ & 60 & 85.88\,(1.81) & 77.96\,(2.29) & 50.79\,(1.57) & 48.94\,(3.82) \\
Shallow-first, $l=6$ & 66 & 85.80\,(1.75) & 78.39\,(2.42) & 50.47\,(1.06) & 48.73\,(4.54) \\
LS-B $k=4$, $K=3$ (default) & 72 & 85.87\,(1.62) & 77.98\,(2.16) & 52.12\,(1.61) & 49.73\,(3.99) \\
Shallow-first, $l=8$ & 84 & 86.11\,(1.77) & 78.34\,(2.29) & 51.84\,(1.40) & 49.49\,(4.41) \\
Full BiLoRA & 120 & 85.68\,(1.60) & 78.23\,(2.04) & 54.64\,(1.87) & 51.43\,(3.48) \\
\midrule
Above hull ($k=4$ / $k=2$) & & $-0.13$\,/\,$+0.02$ & $-0.42$\,/\,$-0.48$ & $+0.56$\,/\,$-0.09$ & $+0.29$\,/\,$+0.08$ \\
Gain over interpolation & & $-0.04$\,/\,$+0.36$ & $-0.40$\,/\,$+0.14$ & $+1.20$\,/\,$+0.74$ & $+0.75$\,/\,$+0.74$ \\
Difference from full BiLoRA & & $+0.19$\,/\,$+0.21$ & $-0.25$\,/\,$-0.27$ & $-2.52$\,/\,$-3.85$ & $-1.70$\,/\,$-2.49$ \\
\bottomrule
\end{tabular}
\end{center}
\end{table}

\paragraph{Accuracy.} On CUB-200 with iBOT and DINO the ImageNet-R pattern recurs: accuracy rises with storage, and LS-B with $k=4$ lies above the upper convex hull of the other configurations (by 0.56 and 0.29~pp) and above the interpolation of the static prefixes (by 1.20 and 0.75~pp), while remaining 2.52 and 1.70~pp below full BiLoRA. On CIFAR-100 and on CUB-200 with AugReg the eight configurations lie within 2.00 and 1.81~pp of each other, against seed ranges of 1.3--2.4~pp; there LS-B with $k=4$ is 0.13 and 0.42~pp below the hull and 0.04 and 0.40~pp below the interpolation, all blocks shared is nominally the best configuration on CUB-200 (78.50\%), and LS-B is 0.19~pp above full BiLoRA on CIFAR-100. Over the sixteen registered hull tests ($k\in\{2,4\}$ in the seven ten-task conditions and at $T=20$), twelve points lie on or above the hull and four below it; the ImageNet-R iBOT point is above it on seeds 0--2 and below it over all 16 seeds (Appendix~\ref{app:results}).

\begin{table}[h]
\caption{Split ImageNet-R with $T=20$ tasks of ten classes, AugReg (seeds 0--2; in parentheses: range). Final: ACC after the last task; Avg: mean of the accuracies after each of the 20 tasks. Storage relative to full BiLoRA (240 slots). Gain: seed-paired gain over the same-seed interpolation of the static shallow-first allocations (pp).}
\label{tab:supp_t20}
\begin{center}
\small
\begin{tabular}{lccccc}
\toprule
Configuration & Slots & Storage & Final & Avg & Gain \\
\midrule
All shared & 12 & 5.0\% & 58.91\,(0.18) & 63.62\,(0.66) & -- \\
Shallow-first, $l=2$ & 50 & 20.8\% & 60.12\,(0.90) & 64.50\,(0.82) & -- \\
LS-B $k=2$, $K=3$ & 80 & 33.3\% & 64.89\,(0.33) & 69.46\,(0.31) & $+3.12$ \\
Shallow-first, $l=4$ & 88 & 36.7\% & 62.22\,(0.07) & 65.79\,(0.34) & -- \\
LS-B $k=4$, $K=3$ (default) & 112 & 46.7\% & 66.02\,(0.95) & 70.18\,(0.64) & $+2.20$ \\
Shallow-first, $l=6$ & 126 & 52.5\% & 64.75\,(0.58) & 67.90\,(0.11) & -- \\
Full BiLoRA & 240 & 100.0\% & 68.43\,(0.93) & 71.87\,(0.34) & -- \\
\bottomrule
\end{tabular}
\end{center}
\end{table}

\paragraph{Twenty tasks.} With $T=20$ (Table~\ref{tab:supp_t20}, Figure~\ref{fig:supp}e), LS-B with $k=4$ retains 112 of 240 slots (47\%, against 60\% at $T=10$, since the ratio approaches $k/L$) and reaches 66.02\%, 2.20~pp above the interpolation of the static prefixes, 0.42~pp above the hull of the other configurations and 2.41~pp below full BiLoRA (1.46~pp at $T=10$); with $k=2$ (80 slots, 33\%) the gain over the interpolation is 3.12~pp. Accuracy averaged over the 20 tasks orders the configurations in the same way. No comparison across $T$ was registered.

\begin{table}[h]
\caption{Registered direction test: both quantities were predicted to order the datasets as ImageNet-R $\geq$ CIFAR-100 $\geq$ CUB-200 on every backbone (pp, seed-paired, seeds 0--2). Observation only: LS-B with $K=1$, which commits blocks 0--3 (56 slots), minus the static blocks 0--3 (48 slots). Gain over interpolation: as in Table~\ref{tab:supp}; on ImageNet-R the $k=4$ point is the default configuration's seeds 0--2. ``--'': no runs (the registered iBOT$\times$CIFAR-100 condition was not run, Appendix~\ref{app:stats}; DINO$\times$CIFAR-100 was not registered).}
\label{tab:supp_order}
\begin{center}
\small
\begin{tabular}{llcccc}
\toprule
Quantity & Backbone & ImageNet-R & CIFAR-100 & CUB-200 & Order holds \\
\midrule
Observation only & AugReg & $+1.66$ & $+0.85$ & $+1.77$ & no \\
 & iBOT & $+1.21$ & -- & $+0.68$ & yes \\
 & DINO & $+1.02$ & -- & $+0.86$ & yes \\
\addlinespace[2pt]
Gain over interpolation, $k=4$ & AugReg & $+0.89$ & $-0.04$ & $-0.40$ & yes \\
 & iBOT & $+0.77$ & -- & $+1.20$ & no \\
 & DINO & $+0.29$ & -- & $+0.75$ & no \\
\addlinespace[2pt]
Secondary: gain, $k=2$ & AugReg & $+1.47$ & $+0.36$ & $+0.14$ & yes \\
 & iBOT & $+1.20$ & -- & $+0.74$ & yes \\
 & DINO & $+0.57$ & -- & $+0.74$ & no \\
\bottomrule
\end{tabular}
\end{center}
\end{table}

\paragraph{Registered direction test.} Before the runs we predicted that the gain from observation alone and the gain over the interpolation both order the datasets as ImageNet-R $\geq$ CIFAR-100 $\geq$ CUB-200 on every backbone. The prediction is not supported: three of the six primary comparisons hold and three are reversed (Table~\ref{tab:supp_order}), and the secondary test with $k=2$ holds on two of three backbones. Observation alone ($K=1$ against the static blocks 0--3, 56 against 48 slots) is positive in all 21 seed pairs of the seven ten-task conditions ($+0.47$ to $+1.95$~pp), but its size does not follow the predicted order.

\paragraph{Committed blocks and cortical signature.} The committed blocks depend on the dataset (Figure~\ref{fig:supp}f). With AugReg, every CIFAR-100 seed commits $\{0,2,3,7\}$ and every CUB-200 seed commits block 11 ($\{6,7,8,11\}$, $\{4,7,8,11\}$, $\{4,6,8,11\}$), so the avoidance of the last block seen on ImageNet-R does not carry over to CUB-200 with AugReg; with iBOT and DINO every CUB-200 seed commits $\{4,6,7,10\}$ and $\{3,5,6,10\}$, and at $T=20$ the AugReg seeds commit $\{2,3,7,8\}$, $\{2,6,8,9\}$ and $\{2,7,8,9\}$. Committed sets differ across seeds in five of the eight conditions (ImageNet-R with every backbone and at $T=20$, CUB-200 with AugReg). The second registered prediction concerned the readout itself: on CUB-200, the share of the task residual of the four top-ranked blocks of the diagnostic readout that V1--V4 carry was predicted to order the backbones as on ImageNet-R and CIFAR-100, iBOT $>$ DINO $>$ AugReg. It holds (75\%, 59\% and 27\%; ImageNet-R 79\%, 42\% and 31\%; CIFAR-100 53\%, 37\% and 22\%, both known when the prediction was registered) and was checked before any CUB-200 training run.

\section{Statistics, registration and deviations}
\label{app:stats}

\paragraph{Registration.} Hypotheses, thresholds and decision rules for the confirmatory comparisons were written down and time-stamped before the corresponding runs were launched; the matched-observation readouts were registered before those runs were launched; the $k$ and $K$ sweeps were added after the confirmatory runs had been judged and are exploratory. The confirmatory family had four criteria per backbone (AugReg, iBOT): a gain of at least 1~pp over random allocation (held); a gain over the same-width controls (held against both activation criteria; the equal-partition control was excluded as a ceiling arm, and a registered response-shuffled control was not run because an offline check found it numerically degenerate); accuracy within 0.42~pp of the best static shallow-first allocation; and backbone-specific placement (both failed: the best static allocation is full BiLoRA on both backbones, 1.46 and 0.47~pp above LS-B). The supplementary conditions (Appendix~\ref{app:supp}) were registered before their runs as exploratory analyses with two directional predictions. Decision rules that require the point estimate to cross a threshold have at most 50\% power when the true effect equals the threshold. With $n$ paired differences, the smallest one-sided sign-test $p$ is $2^{-n}$ (0.125 at $n=3$, 0.0039 at $n=8$), above the corrected level, so non-parametric tests serve only as robustness checks. For the rank analyses of \S\ref{sec:neuro}, $\alpha=.0125$ is a per-backbone level; over two backbones the family-wise error rate is about .0248. The 0.42~pp yardstick is one pair of AugReg runs with $k=6$, measured before a fix to the random-number handling of an earlier selector.

\paragraph{Seeds.} For iBOT, the mean of LS-B's seeds 0--2 (69.60\%) equals the mean of its three best seeds out of 16 (full mean 68.88\%). Seeds were fixed by index before the runs, and we report both values wherever this matters.

\paragraph{Deviations.} Four DINO runs (full BiLoRA seed 2, LS-B $k=2$ seed 2, shallow-first $l=6$ seed 2, and one reproduction of the default configuration) were launched from a working tree with uncommitted changes; we keep them and report the deviation. DINO results for the default configuration use its eight regular runs. Several comparisons span code versions: the default configuration and the activation and equal-partition controls were run under different versions, each of the eight windows mixes two versions with identical allocations and gate records, and the deepest-ten comparison spans two versions. Eighteen early runs (all shared, full BiLoRA and random allocation, seeds 0--2 on AugReg and iBOT) predate the recording of the code version in the run files, so the static references mix runs with and without this record. The registered iBOT$\times$CIFAR-100 supplementary condition (15 runs) was not run: its gate required the 91 runs registered with it to have finished by 2026-09-19 04:00 UTC, and 82 had.

\paragraph{Determinism.} Seven reruns of the default configuration through the code path of the sweep (seeds 0--2 on AugReg and iBOT, seed 0 on DINO) and six reruns of random allocations with explicitly fixed allocation seeds (seeds 0--2 on AugReg and iBOT) reproduced the original final accuracies bitwise.

\end{document}

%% file: math_commands.tex
\usepackage{amsmath,amsfonts,bm}

\def\eqref#1{equation~\ref{#1}}

\def\1{\bm{1}}

\DeclareMathAlphabet{\mathsfit}{\encodingdefault}{\sfdefault}{m}{sl}
\SetMathAlphabet{\mathsfit}{bold}{\encodingdefault}{\sfdefault}{bx}{n}

